\documentclass[11pt]{article}

\usepackage[margin=1in]{geometry}
\usepackage[T1]{fontenc}
\usepackage{newtxtext,newtxmath}
\usepackage{microtype}
\usepackage{amsmath}
\usepackage{graphicx}
\usepackage{booktabs}
\usepackage{tabularx}
\usepackage{array}
\usepackage{enumitem}
\usepackage[font=small,labelfont=bf,skip=6pt]{caption}
\usepackage{xcolor}
\usepackage{fancyhdr}
\usepackage{cite}
\usepackage[hidelinks]{hyperref}
\hypersetup{colorlinks=true,linkcolor=blue!50!black,citecolor=blue!50!black,urlcolor=blue!50!black}

\setlist{itemsep=2pt,topsep=4pt}
\newcolumntype{Y}{>{\raggedright\arraybackslash}X}
\newcommand{\repo}{\url{https://github.com/JordyKieto/puffermoba-dyna}}

\fancypagestyle{firstpage}{\fancyhf{}\fancyfoot[C]{\small\thepage}}

\title{\LARGE\bfseries Learning in Dreams, Winning in Reality:\\ A Continuous Dyna Loop for a Ten-Hero MOBA}
\author{%
  \textbf{Jordy Kieto}\\[4pt]
  \small Code, world models, policies, evaluation and videos: \repo\\
  \small Figures notebook: \href{https://github.com/JordyKieto/puffermoba-dyna/blob/main/docs/moba/paper/figures.ipynb}{\texttt{docs/moba/paper/figures.ipynb}}\\
  \small Companion reports: \href{https://github.com/JordyKieto/carla-imitation-learning}{\textit{Imitation Learning for Autonomous Driving in CARLA}};\\
  \small \href{https://github.com/JordyKieto/neural-simulators}{\textit{Neural Simulator: Lightweight World Models for Continuous Interactive Environments}}
}
\date{}

\begin{document}

\maketitle
\thispagestyle{firstpage}

\begin{abstract}
\noindent World models are usually judged from the inside: by their prediction loss, by the return a policy earns in imagination, or by how convincing their frames look. We judge one from the outside. We learn a structured, multi-agent world model of a complete ten-hero MOBA (206 units, every hero acting every tick, games of up to 6,000 ticks), train a policy only inside it with 1,400-tick free-running imagined episodes, and measure that policy in the real game against the opponent the game ships with. The real game never provides a gradient. It provides two things: the policy's own games as training data for the world model, and an online evaluation that selects and anchors the policy. With these, run as a continuous asynchronous Dyna loop, the policy wins 70.2\% of real games as radiant (421 of 600; 95\% CI 66.4--73.7) on seeds never used for any decision, up from 0\% for dream training alone and 33.7\% before the loop. It wins none as dire, and neither does the shipped opponent when it plays itself. Four findings explain the result. Model exploitation is invisible from inside the dream: every unanchored run collapsed within a few updates while no in-dream metric tracked the collapse. A world model that is accurate on its training corpus is badly wrong on the policy's own games, and Dyna repairs it there, which is worth $+9.2$ points of real win rate with the policy recipe held fixed. Finally, the policy inherits its world model's fidelity profile mechanic by mechanic: the model represents the macro game and not crowd control, cast timing or lethality, and the policy wins by map-wide pressure with almost no coordinated fighting.
\end{abstract}

\noindent\textbf{TL;DR.} A policy trained only in a learned simulator of a ten-hero MOBA wins most of its real radiant games, but only once real data repairs the simulator where the policy goes and real games keep the policy honest; what it learns, and fails to learn, follows the simulator's fidelity, mechanic by mechanic.

\begin{figure}[!htbp]
  \centering
  \includegraphics[width=\linewidth]{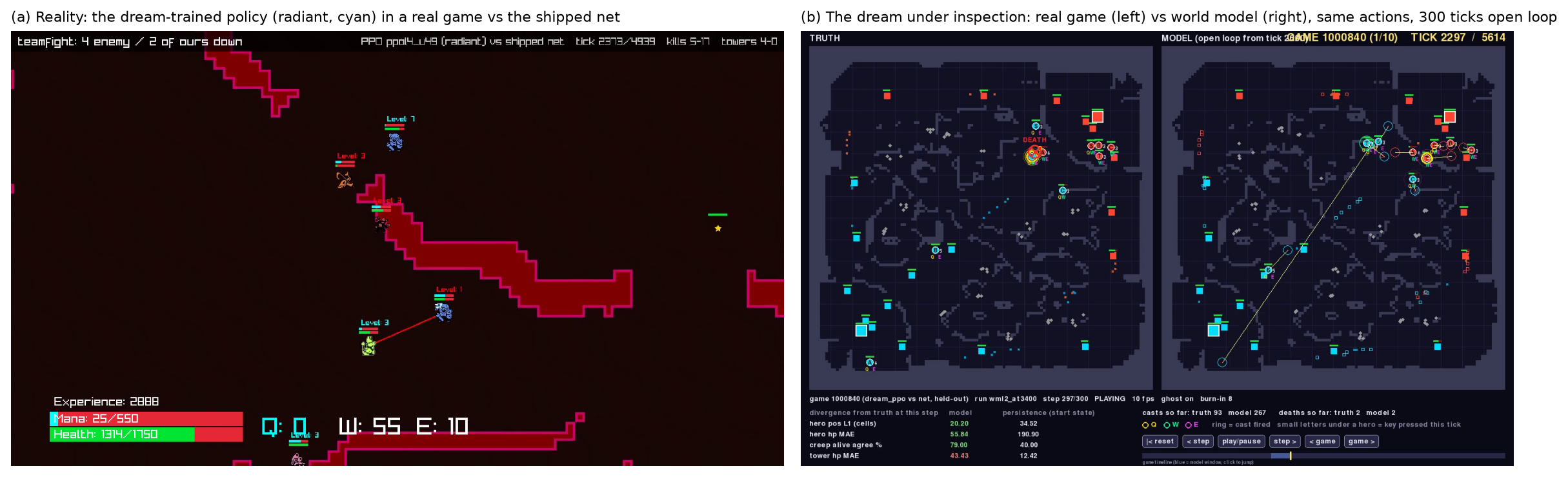}
  \caption{\textbf{(a)} The policy, trained only in the world model, in a real game against the shipped PufferNet, drawn by the game's own client. \textbf{(b)} The world-model debugger: the real game and the dream start from the same state and receive the same actions of all ten heroes; after 300 ticks of open loop, lines join each predicted hero to its true position. The paper is about the gap between these two panels.}
  \label{fig:teaser}
\end{figure}

% ============================================================
\section{Introduction}
\label{sec:intro}
% ============================================================

Learned world models now render playable games~\cite{valevski2025gamengen,bruce2024genie,alonso2024diamond} and train agents with a fraction of the real experience model-free methods need~\cite{hafner2025dreamerv3,micheli2023iris,alonso2024diamond}. The evidence behind that progress comes in two forms. Either the model is scored on its own terms (likelihood, frame fidelity, human preference), or an agent is trained with it, usually from short imagined rollouts branched off real states while real experience keeps flowing in. Neither form answers the question a practitioner faces when the real environment is slow, expensive or unsafe: \emph{if a policy learns by long, free-running imagination in a learned model, what does it actually learn, and does it survive contact with reality?}

This report is the third in a line of work on that question. The first asked whether a policy can learn to drive from demonstrations~\cite{kieto2026carla}; the second whether a compact learned model can stand in for a simulator under continuous control~\cite{kieto2026neuralsim}. Here we close the loop: learn a world, teach an agent inside it, and judge the agent only in the real world.

We pick a setting that stresses every weakness of a learned simulator. PufferLib's MOBA~\cite{suarez2024pufferlib} has 206 units (heroes, creep waves, neutral camps, towers), fifteen abilities with cooldowns, mana and crowd control, ten agents acting at every tick, and games that last thousands of ticks. Our policy trains on 1,400-tick imagined episodes. That is the regime model-based RL is designed to avoid: latent-imagination agents typically imagine 15 to 20 steps~\cite{hafner2020dreamer,micheli2023iris}, and model-based policy optimisation keeps rollouts short precisely because model error compounds with horizon~\cite{janner2019mbpo,talvitie2017self}. We evaluate the way the question demands: real games, a fixed and strong opponent, seeds never touched by any decision, and confidence intervals.

\paragraph{The story in one paragraph.} Dream training alone wins nothing. Long episodes with a win reward and less reward for what the dream overpays (experience) bring the policy to a third of real radiant games. Then every attempt to train further collapses: real win rate drops to near zero within a handful of updates, at any learning rate we tried, while the dream reports steady progress. The cause is the oldest problem of learning in a model~\cite{ha2018worldmodels}, exploitation, but in a form we could only see from outside: real kills and deaths fall in lockstep with real wins, because the policy is learning to stop fighting, while nothing measured inside the dream tracks it. Two remedies, applied together, turn the collapse into steady improvement. A continuous Dyna loop~\cite{sutton1991dyna} retrains the world model on the policy's newest real games, which repairs the model exactly where the policy goes; and a KL anchor ties the policy to the best real-game policy so far, an anchor that moves only when a candidate wins on fresh seeds. The resulting policy wins 70\% of its real radiant games. When we look at \emph{how} it wins, we find that it plays the game its world model can represent: the macro game of lanes and towers, not the micro game of stuns, timed heals and burst kills.

\paragraph{Findings.} Each is measured (Section~\ref{sec:results}); the last is an inference supported by two independent measurements.
\begin{enumerate}[leftmargin=*]
  \item \textbf{Long free-running imagination can transfer.} A policy trained only on 1,400-tick dreams wins 70.2\% of real radiant games. A ladder of policies evaluated on the same untouched games attributes the gain to its ingredients: 0\%, 1.3\%, 5.7\%, 33.7\%, 53.0\%, 70.2\%.
  \item \textbf{Exploitation is invisible from inside the dream.} Over the collapse, dream reward, dream experience, dream deaths, policy entropy and update size all fail to track real wins ($|r|\le 0.37$); real kills and deaths track them ($r=0.84$, $0.82$). Online real-game evaluation is part of the method, not only of the reporting.
  \item \textbf{A world model is wrong where the policy goes, and Dyna repairs it there.} On the policy's own games, the pre-Dyna model predicts 1.79$\times$ the real number of tower falls (it is well calibrated on its corpus) and is worse than predicting no change for hero and tower health. After Dyna, tower falls are within 16\% and hero position error after 200 ticks drops 2.4$\times$. With the policy recipe held fixed, the Dyna-trained model is worth $+9.2$ points [3.5, 14.7] of real win rate.
  \item \textbf{The policy inherits its world model's fidelity profile.} Measured mechanic by mechanic on the policy's games, the model reproduces towers, lanes and aggregate damage, and misses crowd control (17\% of real stuns), cast timing and lethality (heroes die 2.4$\times$ too rarely). The policy's real play mirrors this: it takes more towers than the shipped net, but it groups 9\% of the time (the net 54\%), stuns before 7\% of its kills (26\%) and trades 3.3 deaths per kill.
\end{enumerate}

\paragraph{Contributions beyond the findings.} A \emph{world-model debugger} that treats the learned simulator as an experimental subject: paired real and imagined timelines on identical actions, a field-by-field decomposition of every unit, an event layer, and a batch mode that turns every view into a metric (Section~\ref{sec:debugger}). An \emph{evaluation protocol} for dream-trained policies under repeated online selection, with seed sets whose biases are stated and an untouched holdout for reporting (Section~\ref{sec:eval}). And the system itself, released in full: the world model, the dream-PPO harness with an exact GPU port of the opponent, the asynchronous loop, every policy, log and video, at a whole-project cost of \$382 in attributed cloud spend.

\paragraph{What we do not claim.} We do not claim the policy beats the shipped net at the game: the net wins every one of its own radiant games, in less than half the time. We make no claim about the dire side, where nothing wins. And we make no sample-efficiency claim: the loop played about 46,000 real games, most of them to choose checkpoints.

% ============================================================
\section{Related Work and Positioning}
\label{sec:related}
% ============================================================

\paragraph{Learning inside a learned model.} Dyna~\cite{sutton1991dyna} interleaves real experience, model learning and planning on simulated experience; our system is Dyna with the three processes running asynchronously and at scale. Ha and Schmidhuber~\cite{ha2018worldmodels} trained a controller entirely inside a learned dream and observed it exploiting the dream's errors, which they countered by making the dream noisier. We meet the same exploitation in a much larger multi-agent world, show that it cannot be detected from inside the dream, and counter it with real data and real-game anchoring instead.

\paragraph{Latent imagination and short horizons.} The Dreamer line~\cite{hafner2020dreamer,hafner2025dreamerv3}, IRIS~\cite{micheli2023iris} and DIAMOND~\cite{alonso2024diamond} learn behaviours from imagined rollouts of 15--20 steps started from real states, with real data collected throughout; multi-agent variants on StarCraft micromanagement follow the same design~\cite{egorov2022mamba}. MBPO~\cite{janner2019mbpo} argues for short model rollouts on the grounds that error compounds with horizon, a long-standing observation~\cite{talvitie2017self}. We work in the regime these methods avoid, 1,400-step free-running episodes in a ten-agent world, and report what it takes to make it transfer and what it costs.

\paragraph{Generative interactive games.} GameNGen~\cite{valevski2025gamengen}, Genie~\cite{bruce2024genie}, DIAMOND's game model~\cite{alonso2024diamond} and Oasis~\cite{decart2024oasis} generate playable worlds and are judged by visual quality and human play. Our model is symbolic and full-state, small (0.66\,M parameters), and judged by what an agent trained in it does in the real game.

\paragraph{Model quality and control quality.} The objective mismatch of model-based RL~\cite{lambert2020objective} is that a model's likelihood does not predict the controller's performance; MuZero~\cite{schrittwieser2020muzero} sidesteps reconstruction altogether. We add a mechanistic version of the mismatch: fidelity measured per game mechanic, on the policy's own state distribution, predicts \emph{what kind} of play the policy acquires.

\paragraph{Regularising against exploitation.} Offline model-based RL penalises model uncertainty~\cite{yu2020mopo,kidambi2020morel}; behaviour-\allowbreak regularised RL keeps the policy close to a reference~\cite{wu2019brac}, as RL from human feedback does with a KL term~\cite{ouyang2022rlhf}. Our anchor is such a reference, chosen and moved by real-game performance on fresh seeds: a cheap, online form of the same idea.

\paragraph{MOBA and RTS agents.} OpenAI Five~\cite{berner2019dota} and AlphaStar~\cite{vinyals2019alphastar} reached top human level by training in the real game at enormous scale. We study the opposite corner: no real gradients, a small model and a small budget.

\paragraph{Evaluation practice.} We follow the advice of Henderson et~al.~\cite{henderson2018matters} and Agarwal et~al.~\cite{agarwal2021precipice} on seeds, intervals and selection, and adds untouched seed sets to keep repeated online selection from inflating the reported number.

% ============================================================
\section{Setting}
\label{sec:setting}
% ============================================================

\paragraph{The game and the opponent.} We use the MOBA from PufferLib's Ocean suite~\cite{suarez2024pufferlib}, unmodified (upstream commit \texttt{42f70d69}), wrapped so that every game is deterministic and every state can be captured and restored byte for byte (a restored state continues identically in 99 of 99 tests). Two teams of five heroes (support, assassin, burst, tank, carry) fight over three lanes of creep waves, neutral camps, 22 towers and two ancients; a game ends when an ancient falls, or at 6,000 ticks. Our policy controls all five heroes of one side. The other side is the \textbf{shipped PufferNet}, the policy PufferLib distributes for this game, run from the same code that recorded our data.

\paragraph{Data.} Every real game is stored as a seed plus actions and replayed into dense per-tick state, verified by state hash. Table~\ref{tab:data} lists the corpora. The world model starts from 2,000 games of the shipped net, scripted bots and random play; by the end, 70\% of all real games are the loop's own policy playing.

\begin{table}[!htbp]
\centering\small
\caption{Real-game data. 10,700 games and 234\,GB in total. Ticks are counted where measured.}
\label{tab:data}
\begin{tabularx}{\linewidth}{l r Y}
\toprule
corpus & games & played by \\
\midrule
\texttt{mix1k} & 1,000 (2.66\,M ticks) & the shipped net and scripted bots; 900 train, 100 held out \\
\texttt{mixrand1k} & 1,000 (2.61\,M ticks) & random, net and scripted mixes (off-distribution coverage) \\
\texttt{dyna1}, \texttt{dyna2} & 910 & early dream policies against the net, random play and each other \\
\texttt{dyna3\_r1} & 300 & the best pre-loop policies against the net \\
\texttt{dyna\_live} & 7,490 (27.0\,M ticks) & 205 checkpoints of the loop's policy against the net, alternating sides \\
\bottomrule
\end{tabularx}
\end{table}

\paragraph{Evaluation protocol.}
\label{sec:eval}
A policy is only ever judged in the real game against the shipped net; the world model scores nothing. Because the loop selects checkpoints from real games, we separate the seeds used for selection from the seeds used for reporting (Table~\ref{tab:seeds}). Every win rate in this paper comes from the untouched holdout sets, with 95\% Wilson intervals, and Newcombe intervals for differences. A game that reaches the time limit counts as not won. Evaluation is deterministic: re-running four policies on the holdout reproduced every count exactly. The two selection sets are visibly optimistic, which is why they exist: the pre-loop policy that looked best on the per-update set (45\%) wins 34\% on the holdout, and the final anchor wins 74\% on the gate set and 70\% on the holdout.

\begin{table}[!htbp]
\centering\small
\caption{Seed sets. Only the holdout sets are ever reported as win rates.}
\label{tab:seeds}
\begin{tabularx}{\linewidth}{l l Y}
\toprule
set (seed base) & games & role \\
\midrule
fixed ($5\times10^7$) & 60 per side, after every update & training curves, drift detection, anchor candidates \\
fresh ($6\times10^7$) & 300 per side, per candidate & the anchor gate \\
holdout, holdout2 ($7\times10^7$, $8\times10^7$) & 300 per side each & reporting only; no choice was ever made on them \\
\bottomrule
\end{tabularx}
\end{table}

% ============================================================
\section{Method}
\label{sec:method}
% ============================================================

\subsection{A structured multi-agent world model}

The world model is a token model of the full game state. Each of the 206 units is one token carrying its position, health, mana, experience, ability and attack cooldowns, stun, slow and movement timers, alive and hit flags, and a $9\times9$ patch of nearby walls; a global token carries the phases of the creep-wave, neutral-respawn and rescan clocks. At every tick the actions of all ten heroes enter as inputs. Two pre-LayerNorm transformer blocks mix the 207 tokens with full attention and a learned distance and team bias, a GRU per unit carries memory across ticks, and typed heads predict regression targets, events and classes (0.66\,M parameters in total). We train it the way it will be used: autoregressively on its own predictions, in chunks of 16 ticks over spans of up to 4,096, with loss weights that decay along the rollout, plus a penalty on predicted tower deaths that do not happen, which calibrates tower falls. Training is federated across a fleet of A40 GPUs that average weights every 50 updates~\cite{mcmahan2017fedavg}.

\paragraph{What is not learned.} Unit spawns (creep waves, neutral camps), creep constants, a hero's level from its experience and the statistics that follow from level, counters and value clamps are supplied by hand-coded rules inherited from the first version of the model. Everything dynamic is learned: movement, fights, damage, healing, mana, cooldowns, stuns, experience, deaths with their same-tick respawn, and tower destruction. Every result includes the rules; no rule was added for any result reported here.

\subsection{Dream PPO}

The policy is PufferLib's MOBA policy (an encoder and a six-layer MinGRU), one network shared by the five heroes of a side, acting on the game's own 510-byte egocentric observation. In the dream, observations are rebuilt from imagined state (byte-identical to the game's on real states), and the opponent is the shipped net ported to the GPU (logits within $5\times10^{-6}$ of the original). Our side is drawn at random in each imagined game. Episodes last 1,400 ticks; eight A40s averaging weights after every PPO update reach about 135,000 agent steps per second. The reward is the game's own, reweighted: $\pm100$ for winning or losing, shaping for tower damage and falls, champion kills and siege presence, the game's experience reward halved because the dream overpays for farming, and, in the final configuration, no death penalty.

\subsection{The continuous Dyna loop}

Four processes run concurrently and never wait for one another. A world-model fleet trains without stopping, drawing a quarter of every batch from the hundred most recent real games of the current policy and publishing a checkpoint every 200 updates (about seven minutes). A PPO fleet trains without stopping and swaps each new world model in at the same update on every rank. A collector plays the newest policy against the shipped net and streams the recorded games to the world model. An evaluator plays 60 real games per side for every new policy.

\paragraph{Anchoring and drift.} The policy's loss includes $0.3\cdot\mathrm{KL}(\pi_{\text{anchor}}\,\|\,\pi)$, where the anchor is the best real-game policy so far. A policy that matches the anchor on the per-update seeds is played 300 games per side on the fresh seeds, and replaces the anchor if it wins at least nine more radiant games, about one standard error. If three consecutive evaluations fall to half the anchor's wins, every rank restarts from the anchor.

\subsection{The world-model debugger}
\label{sec:debugger}

\begin{figure}[!htbp]
  \centering
  \includegraphics[width=0.92\linewidth]{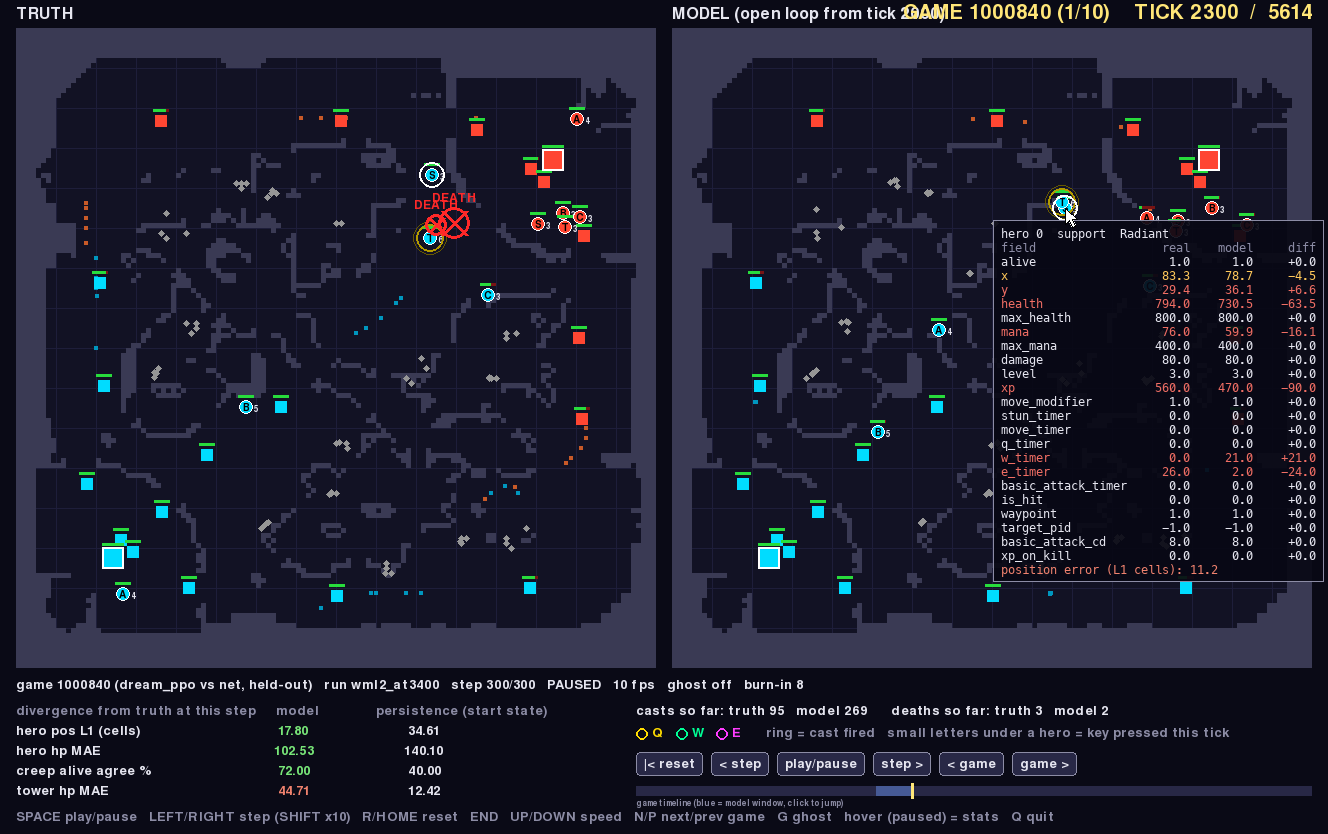}
  \caption{Entity decomposition in the debugger, after 300 ticks of open loop on a real game of the policy (Dyna world model). The radiant support, field by field: the dream has its heal just cast (W cooldown 21; real 0) and has missed the stun it really cast (E cooldown 2; real 26). Over the window the dream fired 269 casts against 95 real ones. The mechanics result of Section~\ref{sec:inherit} in a single tooltip.}
  \label{fig:debugger}
\end{figure}

Aggregate losses say \emph{that} a world model is wrong; they do not say where, when, about whom or about which mechanic, and those are the questions that decide what a policy trained inside it will learn. The debugger is built to answer them, and its design follows from treating the model as an experimental subject (Figure~\ref{fig:debugger}).

The real game and the dream start from the \emph{same} real state and receive the \emph{same} recorded actions of all ten heroes at every tick; after a short warm-up the dream runs open loop on its own predictions, so any difference between the two panels is the model's own accumulated error with the agents' choices held fixed. Every frame of the window is precomputed, so the whole game can be scrubbed, stepped and rewound instantly along a timeline that shows where in the game the model is running; divergence can be traced to the tick where it begins. Any of the 206 units can be decomposed field by field, real against model, with the unit ringed in both panels, so a global error becomes a statement about one unit, one field and one tick. An event layer makes the exploitable mechanics visible: ability casts (detected by cooldown resets), hero deaths (a same-tick jump to base at full health), tower falls, and key presses, which separate a key that was pressed from an ability that actually fired. A ghost overlay draws the true positions on the dream, and every divergence readout sits next to the error of predicting no change. Finally, everything the window shows comes from a programmable replay object, so each view becomes a batch metric over hundreds of windows. Sections~\ref{sec:repair} and~\ref{sec:inherit} are built from that batch mode.

% ============================================================
\section{Results}
\label{sec:results}
% ============================================================

\subsection{The policy wins most real radiant games, and no dire games}

Table~\ref{tab:headline} gives the holdout results. The final anchor wins 213 of 300 radiant games on the holdout and 208 of 300 on the second untouched set, 70.2\% pooled [66.4, 73.7]. That is 37.3 points [29.6, 44.4] above the policy that started the loop and 18.0 points [10.3, 25.4] above the loop's first run. The radiant side alone wins nothing: an untrained policy wins none of its 300 radiant games and takes 0.17 towers per game. Two baselines keep the number honest. The shipped net playing itself wins all 300 of its radiant games, in about 2,270 ticks; our policy wins 70\% and needs about 5,540. And as dire, no policy wins: not ours, and not the shipped net against itself. Our dire play is barely better than random, losing in about 1,290 ticks where a random policy lasts 1,090 and the net 2,270.

\begin{table}[!htbp]
\centering\small
\caption{Real games against the shipped net on the holdout seeds (300 per side). Game length in ticks.}
\label{tab:headline}
\begin{tabular}{l r r r r}
\toprule
policy & radiant wins & dire wins & towers / game & length (radiant / dire) \\
\midrule
untrained policy & 0 / 300 & 0 / 300 & 0.17 & 6,000 / 1,087 \\
shipped net (against itself) & 300 / 300 & 0 / 300 & 4.39 & 2,269 / 2,269 \\
\midrule
start of the loop (E38) & 101 / 300 & 0 / 300 & 3.37 & 5,744 / 1,230 \\
loop, first run & 159 / 300 & 0 / 300 & 3.77 & 5,652 / 1,220 \\
loop with anchor, update 49 & 203 / 300 & 0 / 300 & 3.90 & 5,486 / 1,270 \\
\textbf{loop with anchor, update 69} & \textbf{213 / 300} & 0 / 300 & 3.97 & 5,542 / 1,290 \\
\quad holdout2 & \textbf{208 / 300} & 0 / 300 & -- & 5,537 / 1,287 \\
\bottomrule
\end{tabular}
\end{table}

\subsection{Exploitation is invisible from inside the dream}
\label{sec:collapse}

\begin{figure}[!htbp]
  \centering
  \includegraphics[width=\linewidth]{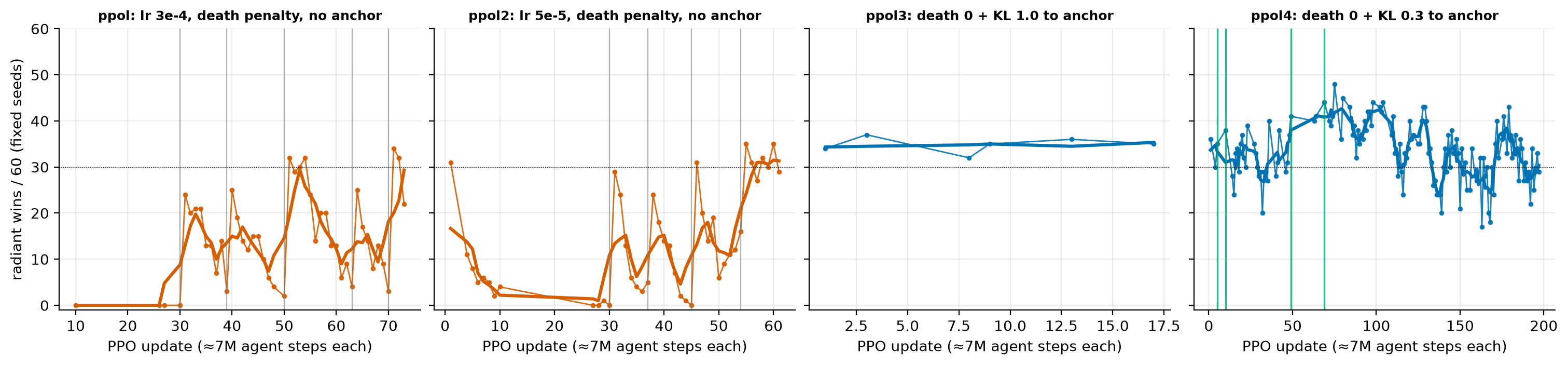}
  \caption{Real radiant wins out of 60 after every PPO update. Without an anchor the policy collapses repeatedly (grey lines: drift resets); with the death penalty removed and a KL anchor it holds and improves (green lines: a new anchor confirmed on fresh seeds).}
  \label{fig:curves}
\end{figure}

\begin{figure}[!htbp]
  \centering
  \includegraphics[width=\linewidth]{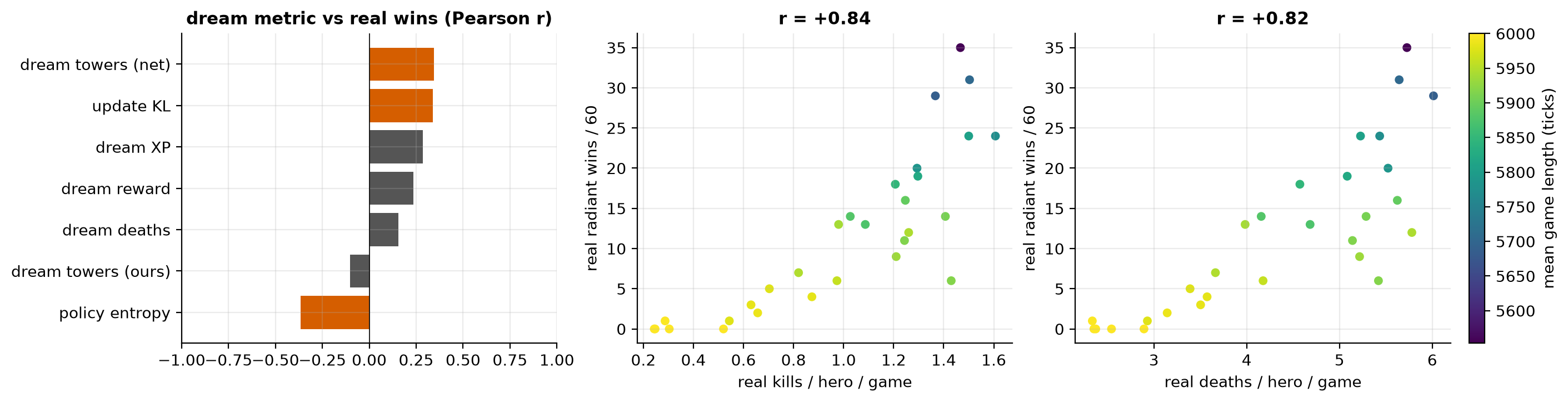}
  \caption{During a collapse, nothing measured in the dream tracks real wins (left), while the policy's real behaviour does: real kills (middle) and deaths (right) per hero and game against real wins, coloured by game length.}
  \label{fig:drift}
\end{figure}

Before the anchor, training could not be continued. Two runs of the loop, at learning rates six-fold apart, rose to about 30 wins in 60 and then fell to between zero and eight within four to six updates, again and again, needing nine resets between them (Figure~\ref{fig:curves}). Lowering the learning rate shrank each update threefold and changed nothing about how many updates the collapse took. Over the collapse of the slower run, no quantity measured inside the dream tracked real performance (Figure~\ref{fig:drift}, left): dream reward ($r=0.23$), dream experience ($0.29$), dream deaths ($0.16$), our dream towers ($-0.10$), policy entropy ($-0.37$). The real game told a clear story. Real kills and real deaths fell together with real wins ($r=0.84$ and $0.82$), and the games stretched to the time limit: the policy was learning to stop fighting. The dream's death penalty, together with its overpayment for safe experience, made disengagement look good in imagination. An ablation from the same starting policy separates the remedies: removing the entropy bonus changes nothing, removing the death penalty delays the collapse by about ten updates, and a KL anchor to the starting policy prevents it. With both changes in the loop, the run never needed a reset and moved its anchor four times in 69 updates, from 53\% to 70\% on the holdout.

\subsection{The world model is wrong where the policy goes, and Dyna repairs it there}
\label{sec:repair}

\begin{figure}[!htbp]
  \centering
  \includegraphics[width=\linewidth]{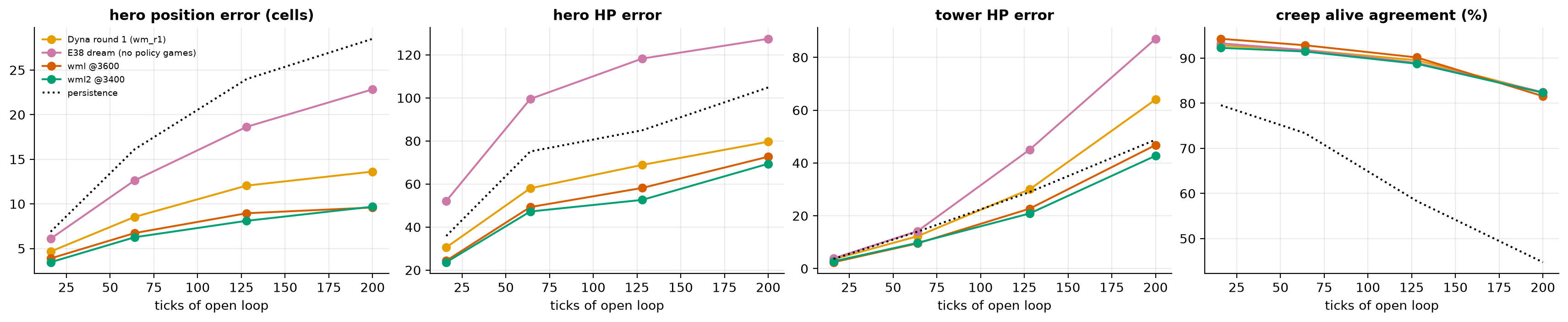}
  \caption{Open-loop divergence from the real game on the policy's own games, by horizon. The pre-Dyna model (pink) is worse than predicting no change (dotted) for hero and tower health; the Dyna models are better at every horizon.}
  \label{fig:divergence}
\end{figure}

On its training corpus the pre-Dyna world model is a good model of towers: 71\% precision on tower falls, with the predicted rate matching the real one. On the policy's own games it is a poor one. We collected 100 real games of the final policy after every model we compare had been saved, so none trained on them, and ran each model open loop from many points in those games. The pre-Dyna model predicts 1.79 times the real number of tower falls at 29\% precision, and after 200 ticks its errors in hero and tower health exceed those of simply predicting that nothing changes (Table~\ref{tab:repair}, Figure~\ref{fig:divergence}). This is the distribution the policy optimises against, and it is wrong in the policy's favour. Training on the policy's newest games repairs it: the Dyna world model predicts 1.16 times the real number of tower falls at 52\% precision and cuts hero position error at 200 ticks from 22.8 to 9.7 cells.

\begin{table}[!htbp]
\centering\small
\caption{World models on the final policy's own real games (unseen by every model). Open-loop error after 200 ticks (300 windows) and tower falls over 128-tick windows (5,265 windows).}
\label{tab:repair}
\begin{tabular}{l r r r r r}
\toprule
world model & hero position & hero HP & tower HP & tower precision & predicted / real falls \\
\midrule
pre-Dyna & 22.8 & 127.5 & 87.0 & 28.6\% & 1.79 \\
Dyna, first round & 13.6 & 79.6 & 64.0 & 39.8\% & 1.27 \\
\textbf{Dyna, continuous} & \textbf{9.7} & \textbf{69.4} & \textbf{42.7} & \textbf{51.8\%} & \textbf{1.16} \\
predict no change & 28.5 & 104.9 & 48.8 & -- & -- \\
\bottomrule
\end{tabular}
\end{table}

\paragraph{What the repair is worth.} To isolate the world model from everything else the loop changed, we froze it and trained the same recipe from the same start (the loop's first-run policy, 53\% on the holdout) for 40 updates, twice per model with different seeds. In the pre-Dyna dream the policy ends at 49.0\% [45.0, 53.0], slightly below where it started; in the Dyna-trained dream it ends at 58.2\% [54.2, 62.0]. The world model alone is worth 9.2 points [3.5, 14.7] (Table~\ref{tab:ladder}, right).

\subsection{What each ingredient added}

Table~\ref{tab:ladder} (left) evaluates one representative policy per stage of the project on the same holdout games. The rungs are separate experiments with their own seeds and lengths, so a rung's step is what a change added in practice rather than a matched estimate; the matched estimate for the world model is the control on the right. Dream training with the game's reward and short episodes wins nothing. Long episodes with a win reward produce the first wins, and the largest single step comes from halving the reward for experience, which the dream overpays. The Dyna loop and the anchor account for the rest.

\begin{table}[!htbp]
\centering\small
\caption{Left: one change per rung, radiant wins on the holdout (the last rung pools both untouched sets). Right: the frozen-world-model control.}
\label{tab:ladder}
\begin{minipage}{0.55\linewidth}\centering
\begin{tabular}{l r}
\toprule
rung & radiant \\
\midrule
untrained policy & 0.0\% \\
dream PPO, game reward, short episodes & 0.0\% \\
+ long episodes, win reward & 1.3\% \\
+ a PPO fleet & 5.7\% \\
+ experience reward halved, shaping & 33.7\% \\
+ continuous Dyna loop & 53.0\% \\
+ no death penalty, KL anchor & \textbf{70.2\%} \\
\bottomrule
\end{tabular}
\end{minipage}\hfill
\begin{minipage}{0.42\linewidth}\centering
\begin{tabular}{l r}
\toprule
frozen world model & radiant (2 seeds) \\
\midrule
pre-Dyna & 294 / 600 = 49.0\% \\
Dyna-trained & 349 / 600 = 58.2\% \\
\midrule
difference & $+9.2$ [3.5, 14.7] \\
\bottomrule
\end{tabular}
\end{minipage}
\end{table}

\subsection{How the policy plays}

\begin{figure}[!htbp]
  \centering
  \includegraphics[width=\linewidth]{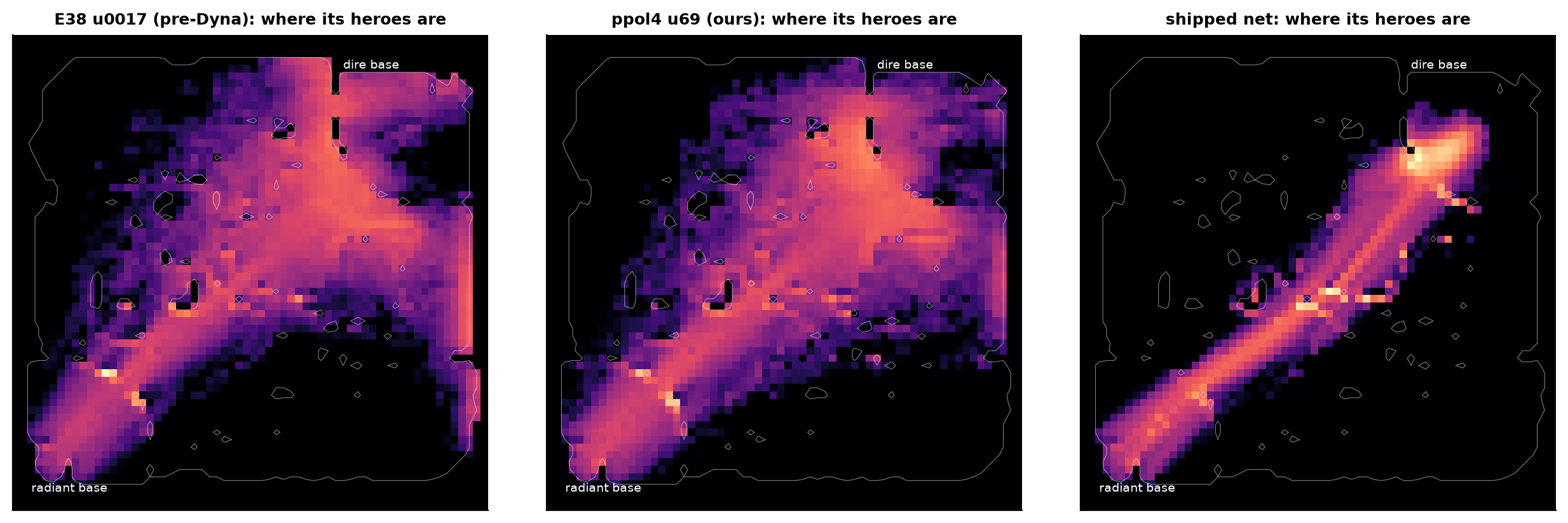}
  \caption{Where the radiant heroes spend their time, over 20 games each: the pre-loop policy (left), the final policy (middle) and the shipped net (right). The net moves as one group down the middle lane; both dream-trained policies cover the whole map.}
  \label{fig:occupancy}
\end{figure}

We played 20 holdout games each with the final policy, the pre-loop policy and the shipped net on the radiant side, and recorded every hero at every tick (Table~\ref{tab:behaviour}, Figure~\ref{fig:occupancy}). The policy does not win by teamwork. Its heroes are within eight cells of two others 9\% of the time; the shipped net's are 54\% of the time. It takes about ten enemy towers per game, more than the net's six, but most fall with fewer than one of our heroes nearby: creeps do the damage while heroes keep up the pressure. It trades lives freely, 3.3 deaths per kill. Its support casts its area heal as often as the net's does, and the heal reaches 0.13 allies per cast against the net's 0.83. Two heroes of ours on the same enemy do focus fire, but they are rarely on the same enemy. The occupancy maps say the rest: the net marches all five heroes down the middle lane and wins in under 2,300 ticks, while our policy spreads over every lane and the jungle and wins by attrition. The loop did not change this strategy; the pre-loop policy plays the same way and converts less of it.

\begin{table}[!htbp]
\centering\small
\caption{Behaviour in real radiant games (20 holdout games each).}
\label{tab:behaviour}
\begin{tabular}{l r r r}
\toprule
 & ours (final) & pre-loop & shipped net \\
\midrule
won & 80\% & 35\% & 100\% \\
game length (ticks) & 5,634 & 5,778 & 2,200 \\
ticks with $\ge$3 heroes within 8 cells & 9\% & 9\% & 54\% \\
our heroes near an enemy tower as it falls & 0.65 & 0.63 & 2.05 \\
enemy towers destroyed & 9.95 & 9.30 & 6.2 \\
kills / deaths per game & 9.4 / 30.7 & 6.9 / 29.1 & 35.7 / 30.0 \\
support heal casts per minute / allies reached per cast & 4.4 / 0.13 & 4.4 / 0.19 & 4.7 / 0.83 \\
enemy deaths preceded by a stun & 7\% & 12\% & 26\% \\
\bottomrule
\end{tabular}
\end{table}

\subsection{The policy inherits its world model's fidelity profile}
\label{sec:inherit}

To see why the policy plays this way, we measured the world model mechanic by mechanic on the policy's own games. Each event (an ability cast, a heal, a displacement such as a blink strike or a hook, the start of a stun or a slow, a hero death, a tower fall) is detected in the real game and in the dream and matched by hero within four ticks, over 64-tick open-loop windows. Table~\ref{tab:inherit} sets the result beside the policy's behaviour.

\begin{table}[!htbp]
\centering\small
\caption{Fidelity of the Dyna world model per mechanic (precision / recall / predicted-to-real count, on the policy's own games), beside the policy's real behaviour.}
\label{tab:inherit}
\begin{tabularx}{\linewidth}{l l Y}
\toprule
mechanic & fidelity in the dream & the policy in the real game \\
\midrule
tower falls & 33\% / 35\% / 1.06 & takes 9.95 towers per game (net 6.2) \\
lanes, waves, positions & waves by rule; positions within $\approx$10 cells at 200 ticks & covers every lane, pushes with the waves \\
hero deaths & 47\% / 20\% / 0.42 & over-extends: 3.3 deaths per kill \\
stuns, slows & 2\% / 0\% / 0.17;\; 3\% / 1\% / 0.34 & a stun precedes 7\% of its kills (net 26\%) \\
heals & 3\% / 3\% / 0.91 & reaches 0.13 allies per heal (net 0.83) \\
ability casts & 24\% / 48\% / 2.02 & casts at the net's rate, rarely in combination \\
\bottomrule
\end{tabularx}
\end{table}

The pattern is one to one. What the dream represents well, the policy has learned and uses: towers fall at the right rate in the dream, and the policy is a better tower-taker than the net. What the dream lacks, the policy does not do: the dream has almost no stuns or slows and heals at the wrong moments, and the policy almost never sets up a kill with a stun or heals an ally in a fight. What the dream distorts, the policy exploits in the direction of the distortion: heroes in the dream survive fights that kill them in reality, and the policy over-extends. Aggregate quality hides all of this; the Dyna model is better on every aggregate measure, and the policy's strategy did not change. Our reading is that a dream-trained policy's competence is bounded by the fidelity of its world model, mechanic by mechanic.

This is an inference from two measurements, not yet an experiment. The test is an intervention: raise the dream's fidelity for crowd control and lethality, or take those mechanics out of the real game, and check that stun-before-kill, ally healing and grouping move with it. Two observations do not fit the simplest version of the story. The policy casts abilities at the net's rate although the dream over-fires them twofold, and the pre-loop policy already played the same macro strategy, which the loop sharpened rather than created.

% ============================================================
\section{Discussion}
\label{sec:discussion}
% ============================================================

\paragraph{Long imagination transfers if reality keeps it honest.} Short rollouts are the standard defence against compounding model error. We trained on 1,400-tick dreams and transferred anyway, but only behind two guardrails that live outside the dream: real data for the model, aimed where the policy goes, and real games that select and anchor the policy. Neither is visible from inside. The practical lesson is that a dream-trained policy needs an outside observer, and that the observer's most informative signal is behavioural.

\paragraph{World models for control should be evaluated where the policy plays, per mechanic.} The pre-Dyna model looked good on its corpus and failed on the policy's games; per-mechanic fidelity on the policy's own data predicted the kind of play the policy would learn, and corpus loss did not. We suggest an event-level fidelity table on the target policy's data, next to a persistence baseline, as a standard report for world models meant for policy learning.

\paragraph{The fidelity profile is also a research agenda.} The path beyond 70\% is concrete: model crowd control, cast timing and lethality well enough that the dream rewards coordinated fights; teach defence, which no dream-trained policy here learned; and make the outside observer cheaper. Real kills and deaths track real wins at $r\approx0.8$, so a behavioural proxy measured on a few games per update could replace most of the 38,000 games the loop spent on selection.

\paragraph{On instruments.} Every mechanistic claim in this paper started as something seen in the debugger and became a number through its batch mode. For learned simulators, being able to ask which unit, which field, which tick and which mechanic is what turns a loss curve into an explanation.

% ============================================================
\section{Limitations}
\label{sec:limits}
% ============================================================

All results are against one opponent, the shipped PufferNet, in a game that is lopsided under it; we claim that a dream-trained policy wins most radiant games, not that it is the better player. The dire side is unsolved. The loop runs are single seeds, and the headline is one policy's score, the best of 13 policies ever evaluated on the holdout; the second untouched set reproduces it (69.3\% against 71.0\%). The matched control isolates the world model's contribution at 9.2 points but not the remaining 12 between that control and the full loop, which differ in a refreshed model, a moving anchor, four times the GPUs and more updates. The world model contains hand-coded rules for spawns and level-derived statistics. The real game was used at scale for selection and data, about 46,000 games or roughly 0.2 billion real ticks against 0.5 billion imagined ones, so this is not a sample-efficiency result. Finally, two infrastructure incidents changed what ran: the first world-model fleet trained on one machine for most of its life after the others ran out of memory, and the collector stopped for two hours on a full disk; both are documented in the repository.

% ============================================================
\section{Reproducibility and Cost}
\label{sec:repro}
% ============================================================

Everything is in the repository (\repo): the deterministic game wrapper and the renderer that records the real client; the world model, dream PPO and arena; the loop, the evaluation service and the scripts that regenerate every number in this paper from the logs; and the figures notebook. Every evaluated policy is stored with its md5. A new anchor is reported by evaluating it on the two holdout sets and re-running two commands, which regenerate every table and figure.

The whole project, including every failed direction, cost \$382 in cloud spend attributed to it by machine identifier (279 machine-hours, 272 of them on GPUs), plus \$214 of unattributed spend on the same account during the same period and about \$100 before billing records begin. The expensive part was the world model, trained early on single H200s; the loop that produced the result ran on A40s at about \$13 per hour and reached its best policy within about four hours of starting.

\section*{Acknowledgements}
We thank the PufferLib authors for the MOBA environment and the shipped policy that served as our opponent.

\bibliographystyle{unsrt}

% ============================================================
\appendix
\section{Additional figures}
% ============================================================

\begin{figure}[!htbp]
  \centering
  \includegraphics[width=\linewidth]{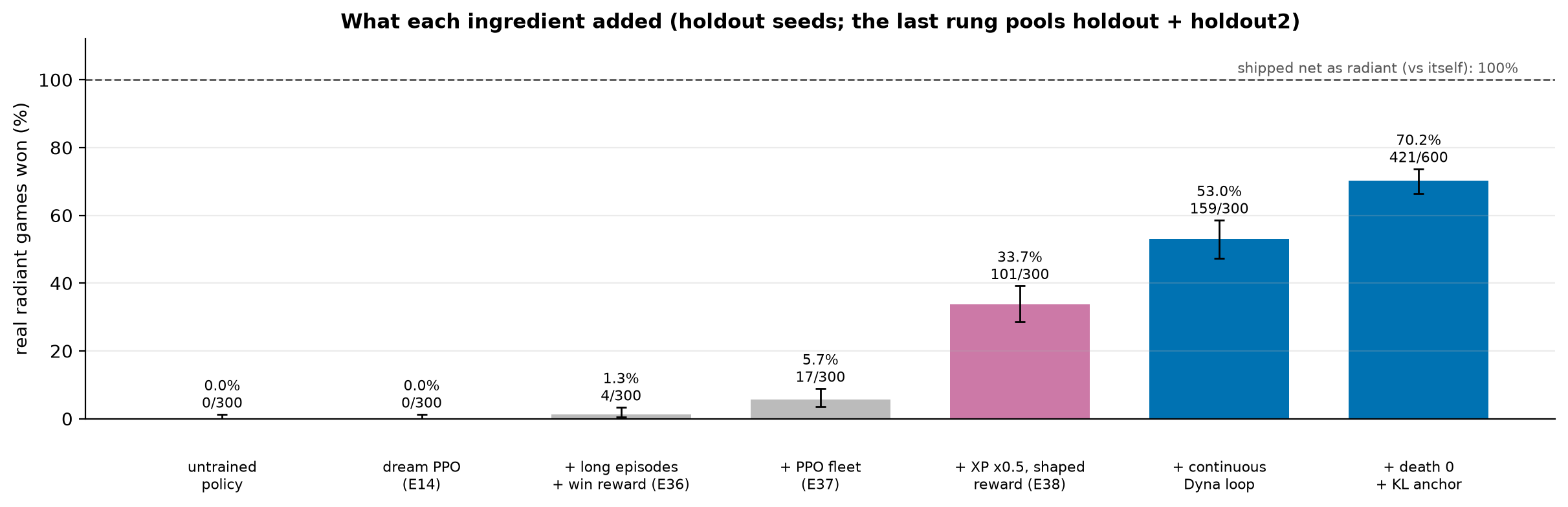}
  \caption{The contribution ladder on the holdout seeds, with 95\% Wilson intervals.}
\end{figure}

\begin{figure}[!htbp]
  \centering
  \includegraphics[width=\linewidth]{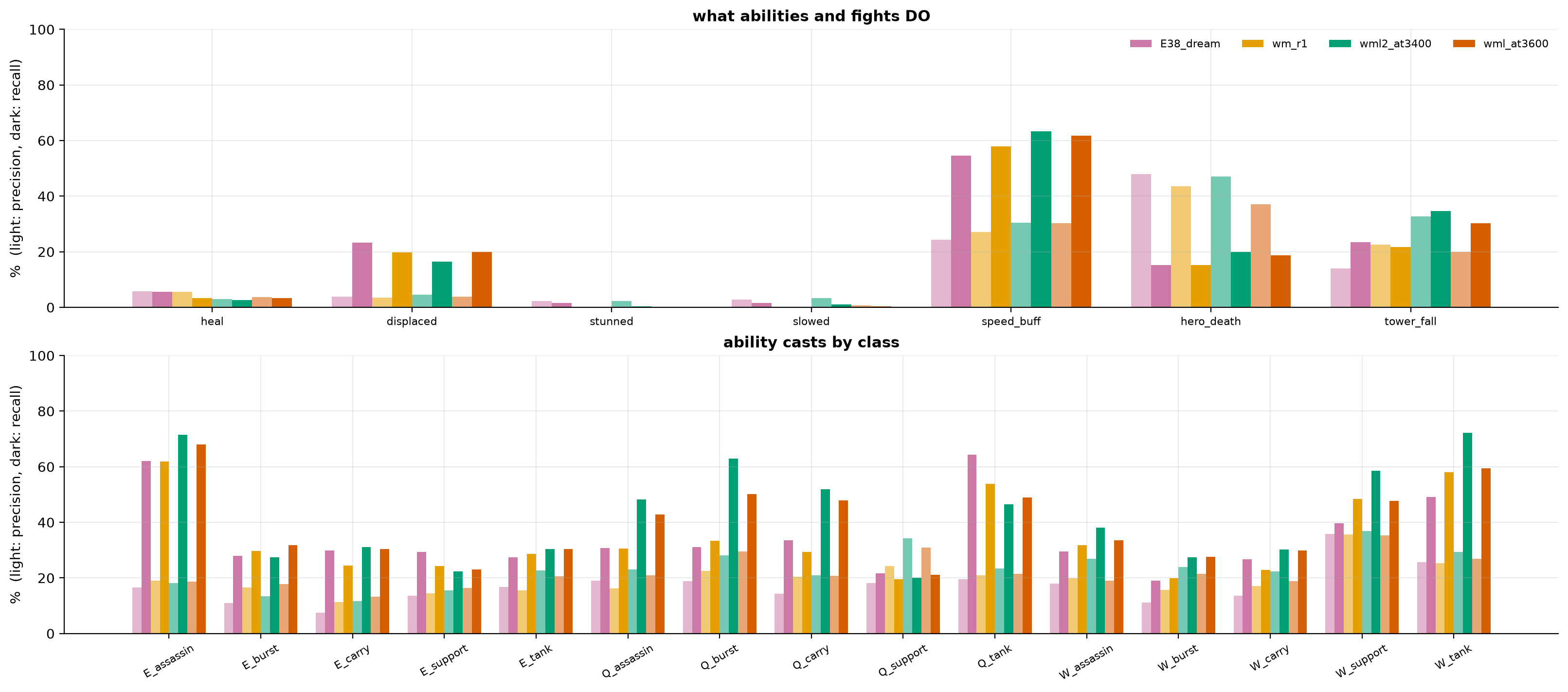}
  \caption{World-model fidelity per mechanic on the policy's own games, for each world-model version (light: precision, dark: recall).}
\end{figure}

\begin{figure}[!htbp]
  \centering
  \includegraphics[width=\linewidth]{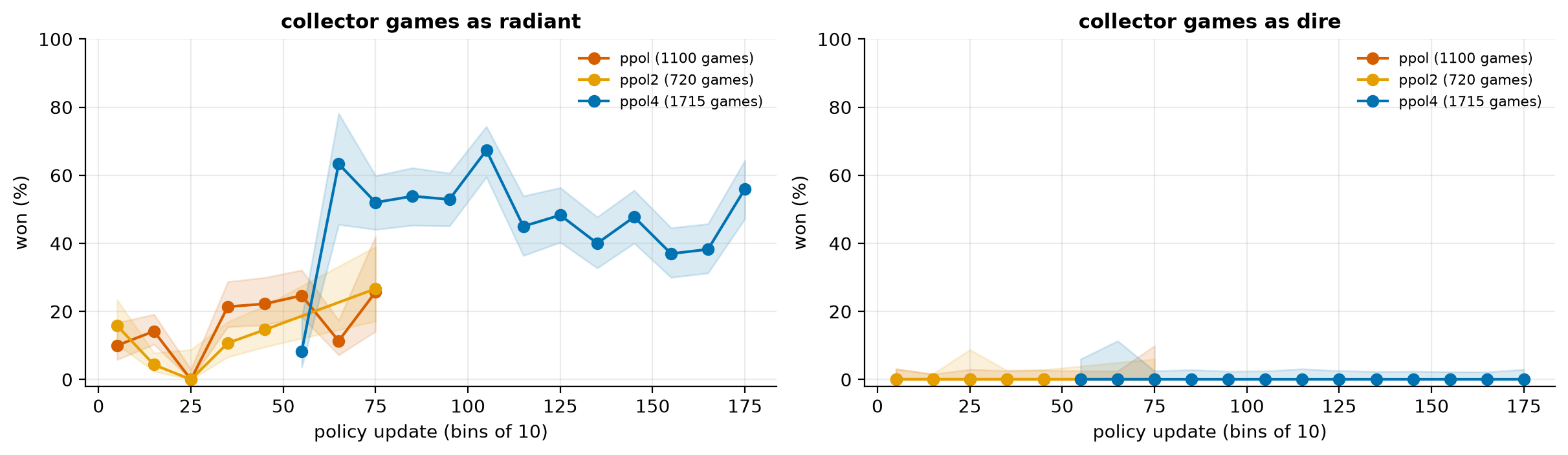}
  \caption{The collector's real games, binned by policy update: an independent stream of real performance on fresh seeds. The anchored run (blue) holds 45--67\% as radiant where the unanchored runs held 0--27\%, and shows a late dip that the drift guard did not catch.}
\end{figure}

\begin{figure}[!htbp]
  \centering
  \includegraphics[width=\linewidth]{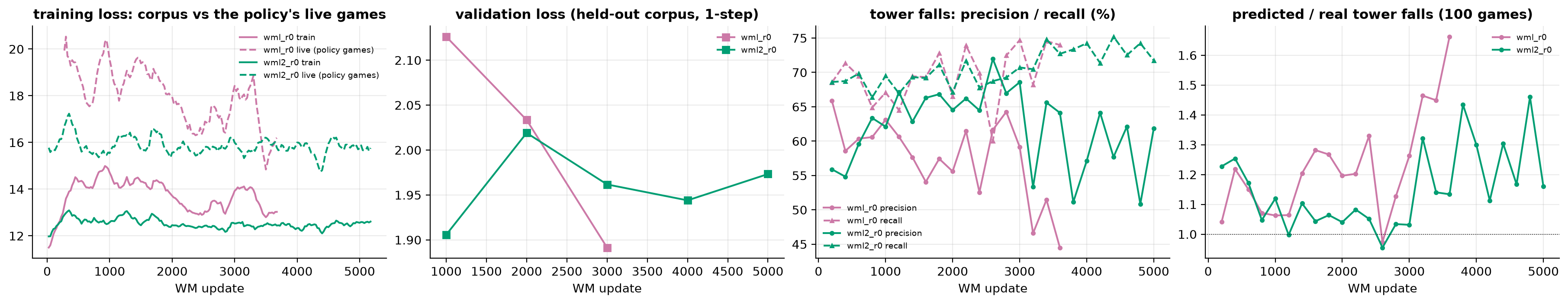}
  \caption{The world model during the loop: training loss on the corpus and on the policy's newest games, held-out validation loss, and tower fidelity per checkpoint.}
\end{figure}

\begin{figure}[!htbp]
  \centering
  \includegraphics[width=\linewidth]{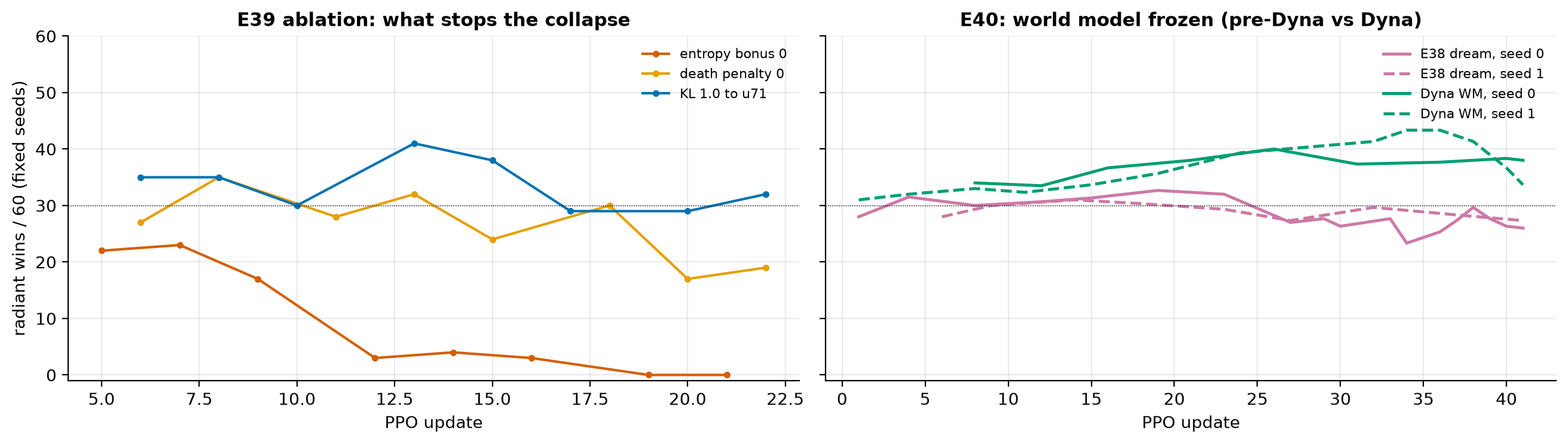}
  \caption{Left: what stops the collapse (one seed per arm). Right: the frozen-world-model control, real wins during training.}
\end{figure}

\end{document}